\documentclass[conference]{IEEEtran}
\IEEEoverridecommandlockouts
\usepackage{makecell}
\usepackage{cite}
\usepackage{amsmath,amssymb,amsfonts}

\usepackage{graphicx}
\usepackage{textcomp}
\usepackage{xcolor}
\usepackage{algorithm}
\usepackage{algorithmic}
\usepackage{preambles}

\usepackage{url}

\makeatletter
\newcommand{\linebreakand}{%
  \end{@IEEEauthorhalign}
  \hfill\mbox{}\par
  \mbox{}\hfill\begin{@IEEEauthorhalign}
}
\makeatother

\usepackage[caption=false,font=footnotesize]{subfig}
\def\BibTeX{{\rm B\kern-.05em{\sc i\kern-.025em b}\kern-.08em
    T\kern-.1667em\lower.7ex\hbox{E}\kern-.125emX}}
\begin{document}
\title{SignRAG: A Retrieval-Augmented System for Scalable Zero-Shot Road Sign Recognition}

\author{
\IEEEauthorblockN{Minghao Zhu}
\IEEEauthorblockA{\textit{Electrical and Computer Engineering}\\
\textit{The Ohio State University;}}
\IEEEauthorblockA{\textit{Applied Research}\\
\textit{Transportation Research Center Inc.}\\
Columbus, Ohio, USA\\
ZhuM@trcpg.com}
\and
\IEEEauthorblockN{Zhihao Zhang}
\IEEEauthorblockA{\textit{Electrical and Computer Engineering}\\
\textit{The Ohio State University}\\
Columbus, Ohio, USA\\
zhang.11606@osu.edu}
\linebreakand
\IEEEauthorblockN{Anmol Sidhu}
\IEEEauthorblockA{\textit{Applied Research}\\
\textit{Transportation Research Center Inc.}\\
East Liberty, Ohio, USA\\
SidhuA@trcpg.com
}
%
\and
\IEEEauthorblockN{Keith A. Redmill}
\IEEEauthorblockA{\textit{Electrical and Computer Engineering}\\
\textit{The Ohio State University}\\
Columbus, Ohio, USA\\
redmill.1@osu.edu}
}

\maketitle
\begin{abstract}
Automated road sign recognition is a critical task for intelligent transportation systems, but traditional deep learning methods struggle with the sheer number of sign classes and the impracticality of creating exhaustive labeled datasets. This paper introduces a novel zero-shot recognition framework that adapts the Retrieval-Augmented Generation (RAG) paradigm to address this challenge. Our method first uses a Vision Language Model (VLM) to generate a textual description of a sign from an input image. This description is used to retrieve a small set of the most relevant sign candidates from a vector database of reference designs. Subsequently, a Large Language Model (LLM) reasons over the retrieved candidates to make a final, fine-grained recognition. We validate this approach on a comprehensive set of 303 regulatory signs from the Ohio MUTCD. Experimental results demonstrate the framework's effectiveness, achieving 95.58\% accuracy on ideal reference images and 82.45\% on challenging real-world road data. This work demonstrates the viability of RAG-based architectures for creating scalable and accurate systems for road sign recognition without task-specific training.
\end{abstract}


\section{Introduction}

Traffic signs are central to safe and efficient transportation. Properly implemented stop signs can reduce relevant intersection crashes by 48.3\% annually \cite{fhwa2009}, and effective ``Wrong Way'' signage significantly lowers wrong-way driving incidents \cite{ntsb2012}. The legibility of these signs also underpins Advanced Driver-Assistance Systems (ADAS) functions such as intelligent speed assistance, which relies on camera-recognized limits to alert drivers \cite{van2014intelligent}.

\begin{figure}[!th]
\centerline{\includegraphics[trim= 2cm 0cm 2cm 0cm,clip,scale=0.2]{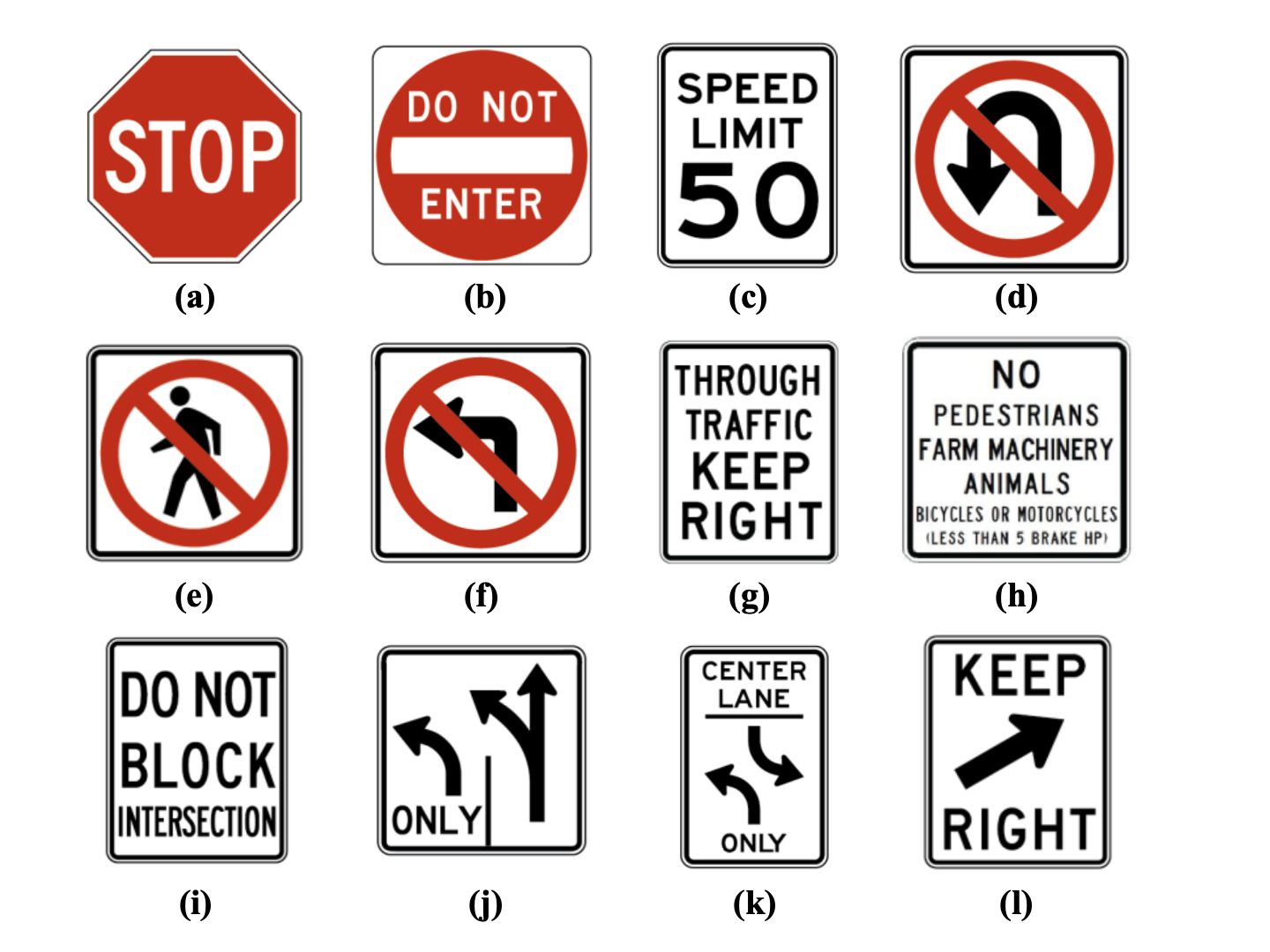}}
\caption{Examples of diverse regulatory signs from the State of Ohio MUTCD catalog used in our experiments, including safety critical signs such as (a) - (c). Other signs may be symbol-based ((d) - (f)), text-based ((g) - (i)), and text and symbol combined ((j) - (l)).}
\label{fig:sign_examples}

\vspace{-0.5cm}

\end{figure}

\begin{figure*}[t]
\centerline{\includegraphics[trim= 0cm 0cm 0cm 0cm,clip,scale=1.0]{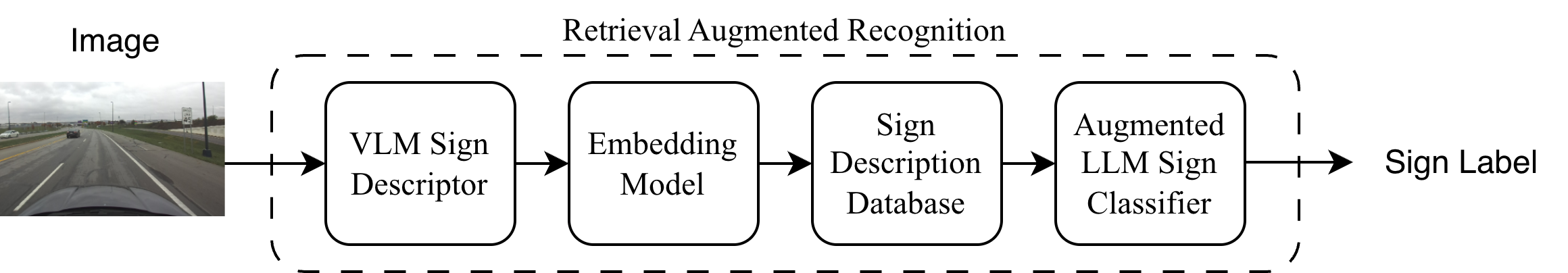}}
\caption{A high-level flowchart of the core components in the proposed method.}
\label{fig:method_flowchart}
\vspace{-0.5cm}
\end{figure*}



Early attempts at automated road sign recognition relied on handcrafted features, using explainable properties like color and shape to identify signs \cite{piccioli1996robust, maldonado2007road, fang2003road}. As computational power increased, the field shifted toward data-driven methods that learn high-dimensional features directly from large-scale datasets. The development of benchmarks like the German Traffic Sign Recognition Benchmark (GTSRB) \cite{stallkamp2012man} and the Tsinghua-Tencent 100K dataset in China \cite{zhu2016traffic} spurred significant progress, leading to high-performance deep learning models. However, replicating this success in the United States has been challenging. While datasets like LISA \cite{mogelmose2015detection} and ARTS \cite{almutairy2019arts} exist, the lack of a single, comprehensive dataset covering the full diversity of signs specified fpr use in the U.S poses challenges for implementing those methods in the U.S. 


The Manual on Uniform Traffic Control Devices (MUTCD) \cite{mutcd2023} defines hundreds of regulatory signs, each with numerous variants. This complexity is compounded by modifications based on jurisdictions, sign installation variability, and real-world degradations. Consequently, building exhaustive, labeled datasets that cover all sign types across various jurisdictions and conditions for traditional supervised deep learning is impractical.

Foundation models offer a promising path to reduce this dependency on task-specific data. However, their direct application can be unreliable, as they are prone to hallucination and operate with fixed knowledge cutoffs. A more robust approach is Retrieval-Augmented Generation (RAG), which grounds the model's reasoning in a reliable, external knowledge base. In a typical RAG pipeline, a retriever fetches relevant information from a corpus, which a generator then uses to produce a contextually grounded output \cite{lewis2020retrieval}.

In this paper, we adapt this paradigm for the task of zero-shot road sign recognition. The high-level workflow of our proposed method is illustrated in Fig. \ref{fig:method_flowchart}. Instead of a text corpus, our knowledge base is a vector database of reference sign designs. Our method first uses a VLM to describe a sign in an image and retrieves the most similar candidates from the database. A LLM then reasons over this retrieved context to determine the final recognition.

The primary contributions of this work are as follows:
\begin{itemize}
    \item A zero-shot recognition framework, capable of recognizing a large and diverse set of road signs, that eliminates the need for task-specific training or fine-tuning, addressing the impracticality of building exhaustive labeled datasets covering all sign classes and conditions.
    \item A novel indexing method that creates a generalized knowledge base. This method uses a VLM to generate abstracted textual descriptions which is crucial for matching real-world variations and makes the system highly scalable.
    \item The use of an LLM as a reasoning-based classifier to perform fine-grained disambiguation of visually similar road signs based on subtle contextual differences.
\end{itemize}

The remainder of this paper is organized as follows. Section II reviews related work in road sign recognition and retrieval-augmented models. Section III details our proposed method, including the indexing, retrieval, augmentation, and generation steps. Section IV presents the experimental setup, datasets, and a thorough analysis of the results in both ideal and real-world scenarios. Finally, Section V concludes the paper and discusses potential directions for future work.

\section{Related Work}

\begin{figure*}[th]
\centering
\includegraphics[trim= 2cm 3cm 2cm 2cm,clip,width=0.95\textwidth]{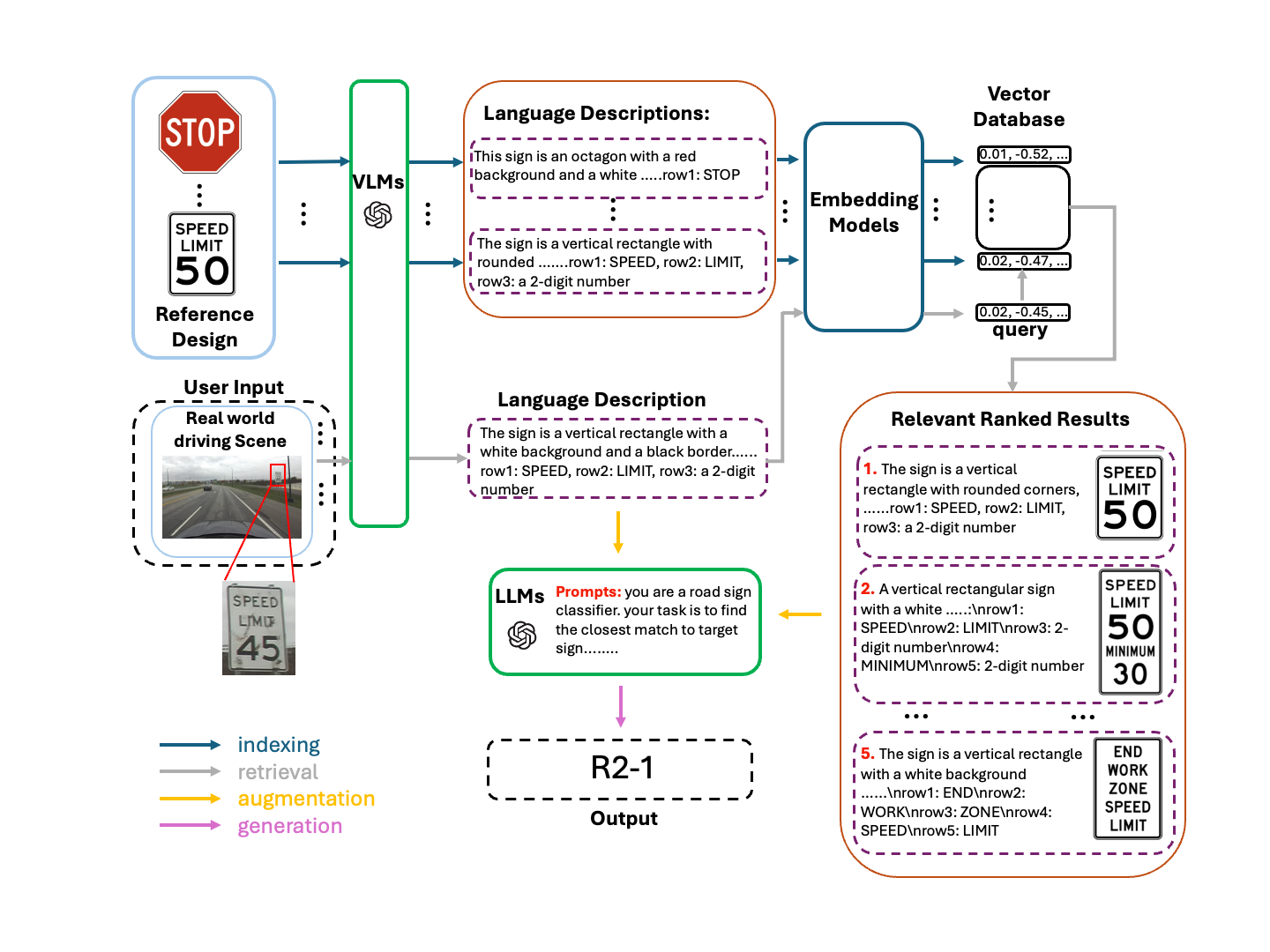}
\caption{The proposed RAG framework for zero-shot road sign recognition. The diagram illustrates both the offline indexing pipeline using reference designs and the online inference pipeline for recognizing a real-world user input. Color coded arrows indicates the four steps.}
\label{fig:stage1_example}
\vspace{-0.5cm}
\end{figure*}

A significant body of research has focused on applying supervised deep learning for automated traffic sign detection. 
A notable example is the work by Almutairy et al. 
\cite{almutairy2019arts}, which established a benchmark to evaluate the efficacy of prominent object detection architectures of the time. 
The authors curated a comprehensive dataset from Vermont interstates, annotated with official MUTCD sign codes, and used it to train and test 
YOLOv3, Faster R-CNN, and RetinaNet-50.
The results underscore the inherent challenges of purely supervised methods. 
The best-performing model, Faster R-CNN, achieved a modest mean Average Precision (mAP-50) of only 67.3\%. The dataset statistics reveal the underlying reason: a severe class imbalance typical of real-world sign inventories. 
While the dataset contained 30 types of regulatory signs, the number of instances ranged from 4,042 images for the common "Speed Limit" sign down to just 25 for the rare "No Parking Anytime" sign. 
The immense effort required to collect and annotate even this limited dataset, coupled with the difficulty of scaling to hundreds of additional rare sign types, highlights the critical need for more data-efficient and scalable paradigms.


Recent work has begun to leverage foundation models to address specific, challenging scenarios where traditional Traffic Sign Recognition (TSR) systems fail. 
For instance, Zhu et al. \cite{zhu2024enhancing} identified a critical failure case for camera-based systems: the recognition of digital variable speed limit (DSL) signs. 
These signs, increasingly used in temporary traffic control zones on highways in states like Ohio and Indiana, employ LED panels whose inherent flickering effect disrupts conventional detectors.
To solve this perception problem, the authors employed the Segment Anything Model (SAM) \cite{kirillov2023segment}, a prominent vision foundation model. They leveraged its powerful zero-shot segmentation capabilities to extract precise pixel masks of the LED panel. 
By aggregating information from these masks across multiple frames, their method achieved an impressive 98.21\% detection rate for this single, challenging sign type. 
This study is a strong proof-of-concept for using foundation models to overcome specific perception hurdles and highlights the power of foundation models.


Beyond perception-specific tasks, researchers are exploring the use of LLMs for high-level reasoning and decision-making in intelligent transportation systems. 
Recognizing the infeasibility of rule-based approaches for every complex driving scenario, Cai et al. \cite{cai2024driving} introduced an LLM-driven agent for making verifiable driving decisions. 
The core of their methodology is a RAG framework. 
This system grounds the LLM's reasoning by dynamically retrieving context from an extensive knowledge base of traffic regulations, driving guidelines, and even court records, which vary significantly across jurisdictions.
By validating their approach in real-world driving, they demonstrated that an LLM can produce a flexible decision-making process based on jurisdiction. Their work establishes a powerful precedent which proves that leveraging RAG to ground an LLM in complex, location-specific regulatory documents is a viable strategy for building trustworthy and transparent transportation systems. 
Our work adapts this core principle to the distinct but related challenge of automated traffic sign evaluation, applying a similar flexible, retrieval-augmented approach to adapt to local official standards.

\section{Method}

Our proposed method is inspired by the RAG framework and, as illustrated in Fig.~\ref{fig:stage1_example}, comprises four key steps: indexing, retrieval, augmentation, and generation. This approach decomposes the complex task of recognizing a road sign from hundreds of potential classes into a more manageable process. The indexing and retrieval steps work in tandem to narrow the search space from an entire sign catalog to a small set of likely candidates. Following this, the augmentation and generation steps use a LLM to perform fine-grained recognition within this reduced candidate pool.

\subsection{Indexing} The foundation of our method is a vector database that serves as a knowledge base of road sign types. This database is constructed in an offline process. First, we use a VLM to generate detailed textual descriptions for a reference image of each sign class (e.g., from the MUTCD catalog of signs). These descriptions are then converted into high-dimensional vector embeddings using a sentence-embedding model and stored in the database. A key aspect of the description generation process is abstraction. To ensure that the embeddings represent the general sign class rather than a specific instance, we instruct the VLM to use generic placeholders for variable content. For example, a speed limit sign showing the number '50' is described as having a "two-digit number," not the specific value '50'. This generalization is crucial for accurately matching real-world signs with variable text or symbols. The resulting database is modular, allowing for the convenient addition or removal of sign classes by simply updating the vector store.

\subsection{Retrieval} When a new real-world driving scene is provided as input, the sign of interest is first processed by the VLM to generate a textual description and a corresponding vector embedding, following the same procedure as in the indexing phase. This query embedding is then used to perform a similarity search against the vector database. We retrieve the top-5 candidates with the highest similarity scores. This practice is analogous to using top-5 accuracy in large-scale classification, a standard method for accommodating model uncertainty between similar classes. This retrieval of multiple candidates, rather than just the single best match, builds resilience against potential inaccuracies or ambiguities in the VLM's initial description. Additionally, since an input image may contain multiple signs, the VLM also generates a textual description of the target sign's location (e.g., "in the upper right quadrant," "to the left of the traffic light"). This natural language localization is crucial for disambiguation, providing the necessary information for any downstream task to process each sign individually and serving as an alternative to bounding boxes.

\subsection{Augmentation} In this step, we construct the final prompt that will be provided to the LLM. The description of the input sign, generated during the retrieval phase, is augmented with the detailed textual descriptions and official sign codes of the top-5 candidates returned by the database search. This process enriches the context, providing the LLM with both the query and a curated set of potential answers.

\subsection{Generation} The augmented prompt is then passed to the LLM, which performs the final recognition. The LLM's task is to reason over the provided context, compare the input sign's features against the candidate descriptions, and identify the most plausible match. Finally, it generates the official sign code of this best match (e.g., "R2-1" for a speed limit sign) as the final output of our method. This step leverages the advanced reasoning capabilities of LLMs to disambiguate between visually similar signs and make an informed decision.


\begin{figure*}[!th]
\centerline{\includegraphics[trim= 2cm 8cm 2cm 0cm,clip,scale=0.35]{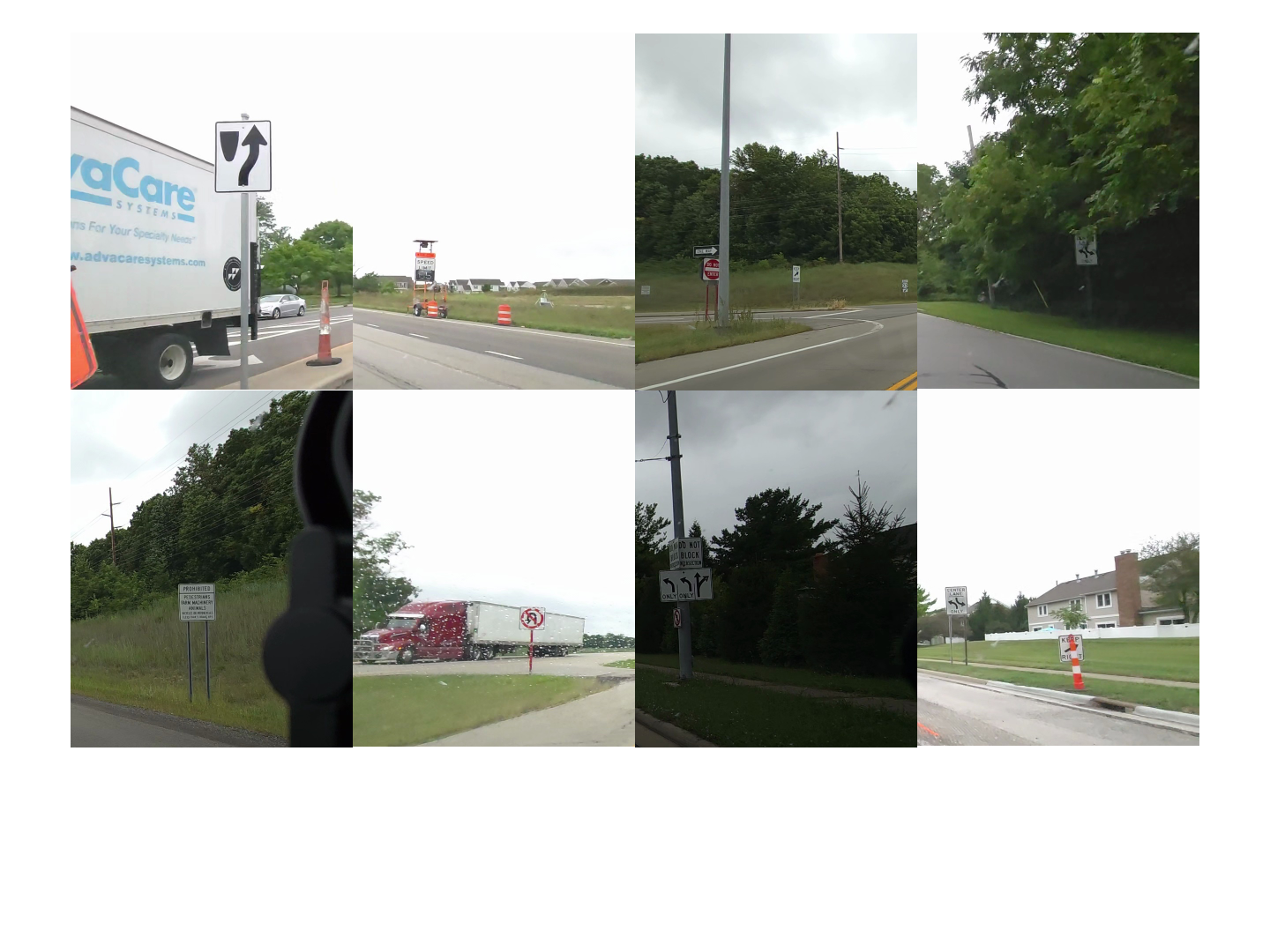}}
\caption{Samples from the real-world dataset, which considers weather, light, and background conditions, sign effectiveness, and rare and emerging sign types.}
\label{fig:realworld_data}
\vspace{-0.5cm}
\end{figure*}

\section{Experiments}

This section details the experimental validation of our proposed method. We design a series of experiments to evaluate its performance on three key aspects: (1) zero-shot recognition capability on a comprehensive set of 303 regulatory signs under ideal conditions; (2) recognition performance in real-world scenarios using on-road data; and (3) the ability to distinguish between in-scope regulatory signs and out-of-scope objects.

\textbf{Experimental Setup}

Our knowledge base is constructed using the 303 regulatory sign designs defined in the Ohio Manual of Uniform Traffic Control Devices (Ohio MUTCD) \cite{omutcd2012}. For real-world evaluation, we use a dataset collected on 72 miles of public roads in Ohio, including local roads, collector roads, and controlled-access highway. The specific models and tools used are as follows:

\begin{itemize}
    \item \textbf{VLM Descriptor:} We employ Gemini 2.5 Flash \cite{comanici2025gemini} to generate textual descriptions. This model was selected for its balance of multimodal reasoning capabilities, inference latency, and cost.
    \item \textbf{Embedding Model:} Textual descriptions are converted to 3072-dimensional vector embeddings using OpenAI's text-embedding-3-large model \cite{openaiemb}.
    \item \textbf{Vector Database:} We use Milvus \cite{wang2021milvus}, an open-source vector database, for its efficient indexing and low-latency similarity search capabilities.
\end{itemize}

\subsection{Experiment 1: Zero-Shot Recognition on Ideal Signs}

First, we evaluate the method's baseline performance and class coverage on the 303 reference sign images from the Ohio MUTCD. Each image features a single, centered sign in perfect condition with a clean background.
The evaluation follows the procedure outlined in the Method section. We define three metrics:

\begin{itemize}
    \item \textbf{Top-1 Accuracy:} The percentage of signs correctly recognized based only on the single best match from the retrieval step.
    \item \textbf{Top-5 Accuracy:} The percentage of signs for which the correct class is present within the top 5 retrieved candidates.
    \item \textbf{Generation Accuracy (Gen Acc):} The final recognition accuracy after the LLM has reasoned over the augmented prompt.
\end{itemize}

To account for model stochasticity, the experiment was repeated five times. The results are presented in Table \ref{tab:sign_det_ideal}.

\begin{table}[h!]
\vspace{-0.3cm}
\centering
\caption{Recognition Performance on Ideal Regulatory Signs in Ohio MUTCD}
\begin{tabular}{||c|c|c|c||} 
 \hline
 \#Run & Top-1 Acc [\%] & Top-5 Acc [\%] & Gen Acc [\%] \\ [0.5ex] 
 \hline\hline
 1 & 88.78 & 99.67 & 95.38 \\ 
 \hline
 2 & 87.13 & 100.00 & 95.71 \\
 \hline
 3 & 89.77 & 99.67 & 96.04 \\
 \hline
 4 & 88.45 & 100.00 & 95.71 \\
 \hline
 5 & 89.11 & 99.67 & 95.05 \\ [0ex] 
 \hline
\end{tabular}
\label{tab:sign_det_ideal}
\vspace{-0.3cm}
\end{table}

Averaged over five runs, the method achieves a mean Top-1 accuracy of 88.65\%. The mean Top-5 accuracy reaches 99.80\%, demonstrating that the retrieval stage is highly effective at capturing the correct class within a small candidate set. The final Generation Accuracy averages 95.58\%, a significant improvement of 6.93 percentage points over the Top-1 baseline. This validates the LLM's role in disambiguating similar candidates and making a more accurate final prediction, all without any task-specific training.

To further justify our decomposed approach, we performed a baseline comparison. We prompted Gemini 2.5 Flash and GPT-5- \cite{openaigpt5} to directly recognize the same 303 sign images in an end-to-end, zero-shot fashion. Despite specifying the scope of signs, both models achieved an accuracy of less than 10\%. This starkly contrasts with our method's 95.58\% accuracy and underscores the necessity of the RAG framework for this complex, many-class task.

\subsection{Experiment 2: Real-World Evaluation}

Next, we assessed the method's performance on our real-world dataset, which contains 181 instances of regulatory signs across 20 types. Sample images from the dataset is shown in Fig. \ref{fig:realworld_data}. These images were captured from a vehicle-mounted camera and feature more challenging conditions, including complex backgrounds, varied lighting, and different viewing angles.
The same models and database from Experiment 1 were used. The results of five runs are shown in Table \ref{tab:real_world_sign_det}.

\begin{table}[h!]
\vspace{-0.3cm}
\centering
\caption{Recognition Performance on Real World Regulatory Signs in Ohio MUTCD}
\begin{tabular}{||c|c|c|c||} 
 \hline
 \#Run & Top-1 Acc [\%] & Top-5 Acc [\%] & Gen Acc [\%] \\ [0.5ex] 
 \hline\hline
 1 & 64.44 & 96.67 & 83.33 \\ 
 \hline
 2 & 61.24 & 96.63 & 84.27 \\
 \hline
 3 & 65.00 & 96.11 & 80.56 \\
 \hline
 4 & 60.77 & 96.13 & 80.66 \\
 \hline
 5 & 65.75 & 96.69 & 83.43 \\ [0ex] 
 \hline
\end{tabular}
\label{tab:real_world_sign_det}
\vspace{-0.3cm}
\end{table}

In this challenging setting, the mean Top-1 accuracy is 63.44\%, while the mean Top-5 accuracy is a robust 96.45\%. The final Generation Accuracy averages 82.45\%. While this represents a performance decrease of approximately 13 percentage points from the ideal setting, it demonstrates that the method remains effective under realistic conditions. The significant gap between Top-1 and Top-5 accuracy highlights the retrieval module's critical role in compensating for visual ambiguities introduced by real-world factors.

\subsection{Distinguishing In-Scope and Out-of-Scope Signs}

A critical capability for a practical system is to reject signs that are not in its target domain (e.g., warning signs, guide signs, or advertisements). We evaluate our method's ability to perform this filtering by analyzing the L2 distance metric from the retrieval step. We process a set of both in-scope and out-of-scope images and plot the distribution of L2 distances for their retrieved candidates.

\begin{figure}[htbp]
\vspace{-0.4cm}
\centerline{\includegraphics[scale=0.35]{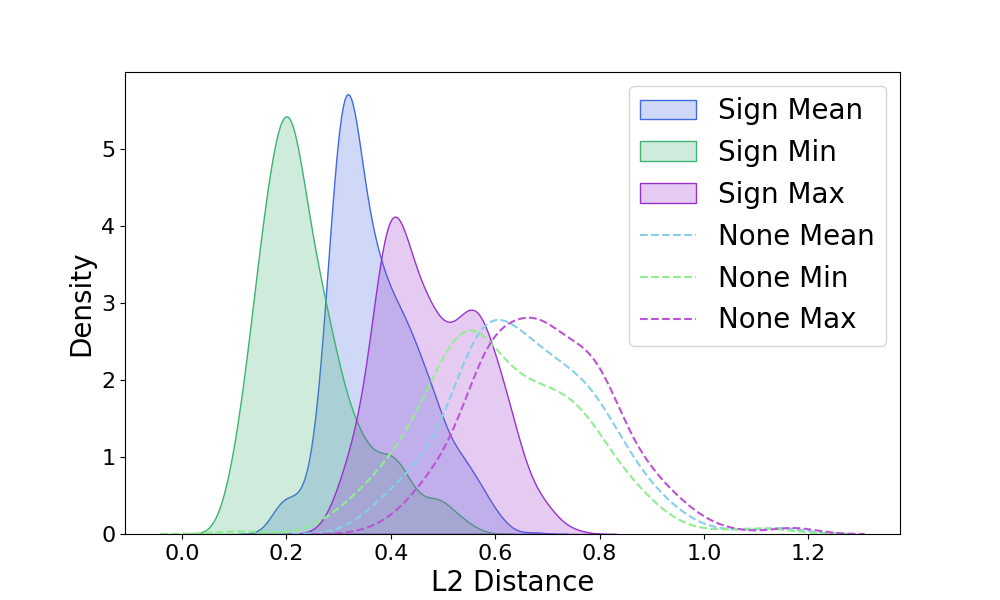}}
\caption{Kernel Density Estimation (KDE) of L2 distances for in-scope ('Sign') and out-of-scope ('None') images.}
\label{fig:unrelated_sign_filter}
\vspace{-0.3cm}
\end{figure}

Figure \ref{fig:unrelated_sign_filter} shows the kernel density estimation (KDE) of these L2 distances. The distributions for in-scope signs are shown as solid fills, while distributions for out-of-scope signs are dashed lines. A clear separation exists between the two groups. This indicates that a simple L2 distance threshold can be used as an effective mechanism to filter out irrelevant signs before the generation step.

Despite these promising results, we acknowledge limitations. Reliance on large foundation models currently restricts real-time application due to inference latency. To quantify this, we benchmarked the pipeline over 100 trials and found the total average latency to be 3.99s, ranging from 2.48s to 32.06s with a standard deviation of 3.32s. A component-level breakdown revealed that the VLM descriptor is the primary bottleneck, averaging 2.48s, followed by the LLM classifier (0.83s) and the database query (0.67s). This latency profile confirms that the cloud-based method is currently impractical for real-time deployment. It is, however, well-suited for offline post-processing, where batch parallelization can improve throughput. Future work toward real-time applications should explore smaller, quantized, or distilled edge-computing favorable foundation models that can be hosted locally to reduce computational and network overhead. Additionally, retrieval speed could be enhanced by optimizing the vector database.

\section{Conclusion}

This paper addressed the significant challenge of recognizing a large and diverse set of road signs in a zero-shot manner, a task where traditional supervised methods are often impractical due to extensive data requirements. We proposed and validated a novel framework inspired by RAG. Our approach effectively decomposes this complex problem by first retrieving a small set of relevant candidates from a knowledge base of reference sign designs, and then leveraging a LLM to perform a final, fine-grained recognition based on reasoned analysis.

Our experimental results demonstrated the efficacy of this method. On a comprehensive set of 303 regulatory signs, the framework achieved 95.58\% accuracy under ideal conditions and a robust 82.45\% on challenging real-world road data—all without any task-specific model training. This performance significantly surpassed that of state-of-the-art end-to-end multimodal models, validating the architectural choice of separating retrieval from generation for this domain.

Despite these promising results, we acknowledge limitations. The reliance on large foundation models currently restricts real-time application due to inference latency. Due to this limitation, currently the method is more suitable for automated sign recognition task in post-processing. Thus, we plan to extend this framework's capabilities beyond simple recognition. The RAG-based structure is well-suited to serve as a specialized perception tool within a larger agentic AI system for road sign evaluation. By grounding the LLM's judgments in official traffic control device manuals, this future work will aim to perform automated sign maintenance checks and verify regulatory compliance, paving the way for a more comprehensive and autonomous infrastructure management system.


\bibliographystyle{IEEEtran}
\bibliography{root}

\end{document}